\documentclass[letterpaper, 10 pt, conference]{ieeeconf}  %

\IEEEoverridecommandlockouts                              %

\usepackage[table]{xcolor}
\usepackage{cite}
\usepackage{textcomp}
\usepackage{stfloats}
\usepackage{url}
\usepackage{verbatim}
\usepackage{graphicx}
\usepackage{times}
\usepackage{paralist}
\usepackage{amssymb}
\usepackage{mathtools}
\usepackage{tikz}
\usepackage{lipsum}
\usepackage{amsmath}
\usepackage{amssymb}
\usepackage{svg}
\usepackage{scalerel}
\usepackage{booktabs}
\usepackage[most]{tcolorbox}

\usetikzlibrary{shadows.blur}

\usetikzlibrary{arrows}
\usetikzlibrary{shapes}
\usetikzlibrary{snakes}
\usetikzlibrary{calc}
\usetikzlibrary{intersections}
\usetikzlibrary{arrows}
\usetikzlibrary{shapes}
\usetikzlibrary{snakes}
\usetikzlibrary{calc}
\usetikzlibrary{tikzmark,fit}
\usetikzlibrary{positioning}
\usetikzlibrary{spy}
\usepackage{pgfplots}
\usetikzlibrary{pgfplots.statistics, pgfplots.colorbrewer, pgfplots.groupplots, pgfplots.fillbetween}
\usepackage{pgfplotstable}
\usepackage{siunitx}
\usepackage{authblk}
\usepackage{multirow}
\usepackage{xspace}

\usepackage{fontawesome5}
\usepackage{xcolor}

\definecolor{pyblue}{HTML}{3776AB} %

\usepackage{algorithm,algpseudocode}
\makeatletter
\let\NAT@parse\undefined
\makeatother
\usepackage[hidelinks]{hyperref}

\usepackage{ifthen}
\usepackage{soul}

\newboolean{authnotes}
\setboolean{authnotes}{true} %

\ifthenelse{\boolean{authnotes}}
{
  \DeclareRobustCommand{\paul}[1]{{\sethlcolor{green!30}\hl{\textbf{Paul}: #1}}}
  \DeclareRobustCommand{\henrik}[1]{{\sethlcolor{blue!30}\hl{\textbf{Henrik}: #1}}}
  \DeclareRobustCommand{\devdutt}[1]{{\sethlcolor{orange!30}\hl{\textbf{Devdutt}: #1}}}
  \DeclareRobustCommand{\todo}[1]{{\sethlcolor{red!30}\hl{\normalfont \textbf{TODO}: #1}}}
  \DeclareRobustCommand{\needref}[1]{\textcolor{red}{[REF #1]}}

}
{
  \DeclareRobustCommand{\paul}[1]{}
  \DeclareRobustCommand{\henrik}[1]{}
  \DeclareRobustCommand{\devdutt}[1]{}
  \DeclareRobustCommand{\todo}[1]{}
  \newcommand{\needref}[1]{}
}

\newcommand{\fakepar}[1]{\vspace{1mm}\noindent\textbf{#1.}}

\makeatletter
\let\MYcaption\@makecaption
\makeatother

\usepackage[font=footnotesize]{subcaption}

\makeatletter
\let\@makecaption\MYcaption
\makeatother

\newcommand{\namingconvention}{paper\xspace} %

\title{\LARGE \bf
The Mini Wheelbot Dataset: High-Fidelity Data for Robot Learning
}

\author{Henrik Hose$^\ast$, Paul Brunzema$^\ast$, Devdutt Subhasish, Sebastian Trimpe%
\thanks{$^\ast$ Equal contribution.}%
\thanks{This work is funded in part by the German Research Foundation (DFG) – RTG 2236/2 (UnRAVeL) and the German Federal Ministry of Research, Technology and Space under the Robotics Institute Germany (RIG).}%
\thanks{All authors are with the Institute for Data Science in Mechanical Engineering (DSME), RWTH Aachen University, Germany (e-mail: \{henrik.hose, brunzema, trimpe\}@dsme.rwth-aachen.de)}%
}

\begin{document}

\maketitle
\thispagestyle{empty}
\pagestyle{empty}

\begin{abstract}
The development of robust learning-based control algorithms for unstable systems requires high-quality, real-world data, yet access to specialized robotic hardware remains a significant barrier for many researchers.
This \namingconvention introduces a comprehensive dynamics dataset for the Mini Wheelbot, an open-source, quasi-symmetric balancing reaction wheel unicycle.
The dataset provides 1~kHz synchronized data encompassing all onboard sensor readings, state estimates, ground-truth poses from a motion capture system, and third-person video logs.
To ensure data diversity, we include experiments across multiple hardware instances and surfaces using various control paradigms, including pseudo-random binary excitation, nonlinear model predictive control, and reinforcement learning agents.
We include several example applications in dynamics model learning, state estimation, and time-series classification to illustrate common robotics algorithms that can be benchmarked on our dataset.
\end{abstract}

\begin{tcolorbox}[
    enhanced,
    colback=white,
    colframe=black!10,
    arc=1pt,
    boxrule=1pt,
    top=2pt,
    bottom=2pt,
    left=2pt,
    right=2pt,
] 
    \footnotesize 
    \hspace*{5em} \faGithub \hspace{0.7em} \href{https://github.com/wheelbot/dataset}{\texttt{github.com/wheelbot/dataset}} \\
    \hspace*{5em} \textcolor{pyblue}{\faPython} \hspace{0.5em} \texttt{pip install wheelbot-dataset}
\end{tcolorbox}

\section{Introduction}

Recent advances in data-driven modeling and learning-based control 
have enabled robotic systems to solve complex control problems
by directly learning control policies from experience \cite{kober2013reinforcement,levine2016end}.
Research in such methods critically depends on diverse, high-quality training data, yet surprisingly few real-world datasets of \emph{fast, unstable, nonlinear, underactuated} robots exist beyond perception-focused vehicles~\cite{kulkarni2023racecar,caesar2020nuscenes} and well-linearizable quadcopters \cite{delmerico2019we}.
In this \namingconvention, we hope to democratize robotics research and enable reproducible benchmarks by introducing a high-fidelity dataset for the recently developed Mini Wheelbot~\cite{hose2025miniwheelbot}:
An open-source, quasi-symmetric balancing reaction wheel unicycle robot.
The Mini Wheelbot balances with a linear state-feedback controller using its driving wheel similar to a segway (pitch) and its reaction wheel for roll stabilization.
However, the yaw angle of the Mini Wheelbot is linearly uncontrollable,
necessitating nonlinear methods like model predictive control (MPC) or reinforcement learning (RL).
The robot can stand up from any initial orientation using its driving and reaction wheels.
This allows for automatic environment resets after failed experiments.
The Mini Wheelbot is designed for experimental ease, featuring a rugged aluminum housing, a \SI{45}{\minute} battery runtime, and a Linux single-board computer (Raspberry Pi CM4) running all controllers onboard.

\begin{figure}[t!]
    \centering
    \includegraphics[width=\linewidth]{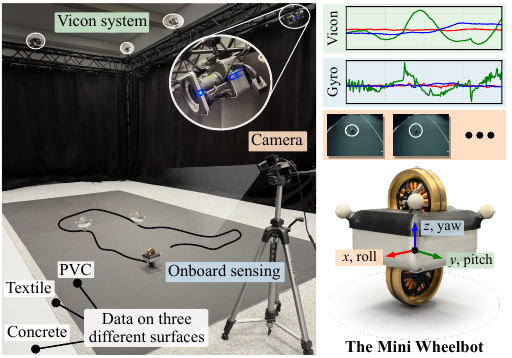}
    \vspace{-1em}
    \caption{Experimental recording setup for the Mini Wheelbot dataset.}
    \label{fig:header}
    \vspace{-1em}
\end{figure}

The contribution of this \namingconvention is a large, high-quality dynamics dataset of the Mini Wheelbot.
The dataset contains~\SI{1}{\kilo\hertz} data of all onboard sensor readings, the estimated state, ground-truth pose measurements from a motion capture system, and third-person view videos of each experiment.
We perform a variety of experiments using pseudo-random binary excitation signals (PRBS) as setpoints of a linear controller, an MPC for driving, and an RL policy that races along tracks.
Experiments are performed across multiple hardware instances and on different surfaces.
With this dataset, we hope to encourage researchers to use the Mini Wheelbot to benchmark their learning-based control methods, even without access to the real hardware.
We include example implementations illustrating the use of the dataset for dynamics learning, state estimation, and time-series classification.

\begin{table}[b!]
\vspace{-1em}
\caption{Experiments in the Mini Wheelbot dataset.}
  \centering
  \label{tab:experiments}
  \resizebox{3.4in}{!}{%
    \setlength{\tabcolsep}{3pt}
    \begin{tabular}{@{}lccccc@{}}
      \toprule
      \textbf{Group} & \textbf{Controller} & \textbf{Reference} & \textbf{\# Trajs.} & \textbf{$\Sigma$ Dur. [min]} & \textbf{\# Crashes} \\
      \midrule
      Pitch              & LQR        & PRBS           & 16              & 6.0                     & 11          \\
      Roll               & LQR        & PRBS           & 45              & 21.1                    & 20          \\
      Vel + Roll         & LQR        & PRBS           & 29              & 2.6                     & 16          \\
      Vel + Pitch        & LQR        & PRBS           & 13              & 1.7                     & 2           \\
      Yaw Random         & AMPC       & PRBS           & 50              & 38.0                    & 5           \\
      Yaw Circles        & AMPC       & Geometric      & 89              & 53.2                    & 4           \\
      Yaw Eight          & AMPC       & Geometric      & 104             & 77.5                    & 14          \\
      Human              & AMPC       & Geometric        & 7               & 13.6                    & 0           \\
      Racetrack          & RL         & Track          & & \makebox[2pt][c]{\textit{(Available soon)}}                      &           \\
      \midrule
      All                &            &                & 383             & 219.7                   & 80          \\
    \end{tabular}%
  }
\end{table}

\section{A Dataset for Dynamics Learning}
We record our dataset with different controllers tracking random and deterministic references to excite all relevant Mini Wheelbot's dynamics.
An overview of the experiments contained in the dataset is given in Tab.~\ref{tab:experiments}: Individual and combined roll and pitch references (with and without driving velocity reference) are recorded using PRBS as standard for system identification.
For these experiments, we use the linear state-feedback controller with decoupled roll and pitch in~\cite{geist2022wheelbot} that does not control the yaw angle.
However, the free system response in yaw is recorded and can be readily used for dynamics modeling.
Experiments based on human direction and velocity commands and geometric references are recorded using a nonlinear MPC that is approximated using a neural network (AMPC)~\cite{hose2024parameter,hose2025miniwheelbot,hose2025fine}.
These experiments exhibit smaller excitation in roll and pitch direction, but represent a state distribution relevant for meaningful tasks such as driving.
We include sequences that lead to a crash as these can contain valuable information right before the crash occurs.
Finally, we record data of the Mini Wheelbot racing along predefined tracks using an RL policy.

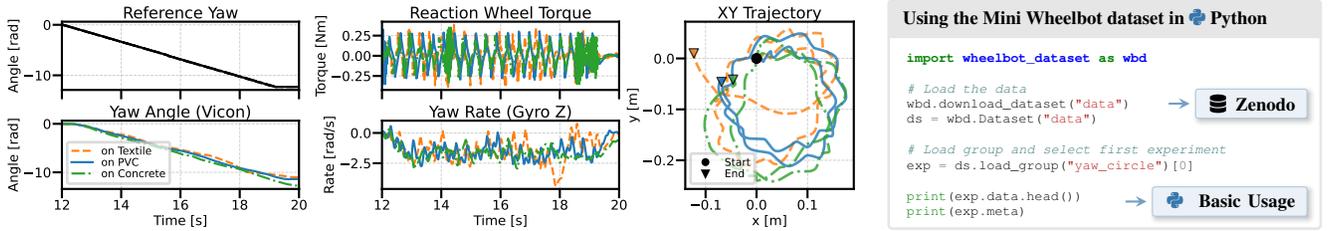
\begin{figure*}[t!]
    \centering
    \begin{minipage}[t]{0.66\linewidth}
        \vspace{0pt} %
        \centering
        \includegraphics[width=\linewidth, height=1.2in, keepaspectratio=false]{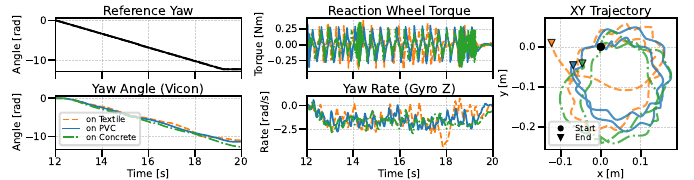}
    \end{minipage}
    \hfill
    \begin{minipage}[t]{0.33\linewidth}
        \begin{tikzpicture}[remember picture, baseline=(codebox.north)]%
            \node[inner sep=0pt, outer sep=0pt, text width=\linewidth, align=left] (codebox) {
                \includegraphics[width=\linewidth]{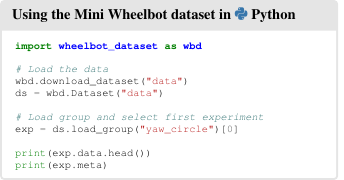}
            };

            \node (linkbtn) [
                anchor=north east, 
                xshift=-8pt, 
                yshift=-71pt, 
                fill=pyblue!10, 
                draw=pyblue!40, 
                rounded corners=1pt, 
                inner sep=2pt,
                align=center,
                blur shadow={ 
                    shadow xshift=1pt, 
                    shadow yshift=-1pt, 
                    shadow blur radius=2pt, 
                    shadow opacity=30 
                },
            ] at (codebox.north east) {
                \href{https://github.com/wheelbot/dataset/blob/main/examples/basic_usage.ipynb}{
                    \scriptsize \textcolor{pyblue}{\faPython} \enspace \textbf{Basic Usage}
                }%
            };

            \node (databtn) [
            anchor=north east, 
            xshift=-8pt, 
            yshift=-34pt, %
            fill=pyblue!10, 
            draw=pyblue!40, 
            rounded corners=1pt, 
            inner sep=2pt,
            align=center,
            blur shadow={shadow xshift=1pt, shadow yshift=-1pt, shadow blur radius=2pt, shadow opacity=30}
            ] at (codebox.north east) {
                \href{https://doi.org/10.5281/zenodo.18234244}{
                    \scriptsize \faDatabase \enspace \textbf{Zenodo}
                }%
            };

            \draw[
                -stealth,          %
                thick,             %
                pyblue!70,         %
            ] ([xshift=-10pt]linkbtn.west) -- ([xshift=-2pt]linkbtn.west);

            \draw[
                -stealth,          %
                thick,             %
                pyblue!70,         %
            ] ([xshift=-10pt]databtn.west) -- ([xshift=-2pt]databtn.west);

        \end{tikzpicture}
    \end{minipage}
    \caption{\textit{Left:} The same reference (Yaw Circles) on the three different surfaces contained in the dataset.
    \textit{Right:} Python snippet on how to load the dataset.}
    \label{fig:placeholder}
\end{figure*}

All data in the dataset is logged at~\SI{1}{\kilo\hertz} directly onboard the robot, thus it is as time-synchronized as the controller would receive it, yet some sensors might provide an updated measurement at a lower rate.
Fields in the comma-separated values format \texttt{.csv} are summarized in Tab.~\ref{tab:dataset_fields}.
All coordinates are aligned with the robot body frame, where the $x$-axis points forward, the $y$-axis sideways and the $z$-axis upwards (see Fig.~\ref{fig:header}).
Metadata fields in JSON (\texttt{.json}) are \texttt{experiment\_status} indicating if the robot crashed, \texttt{wheelbot} which contains the hardware id, \texttt{surface} on which the experiment was conducted, and a unique identifier~\texttt{uuid}. Third-person view videos (\texttt{.mp4}) document how an experiment looks for visual inspection.

\begin{table}[htb]
  \centering
    \caption{Overview of dataset fields and signal semantics.
  }
  \label{tab:dataset_fields}
  \resizebox{3.4in}{!}{%
    \begin{tabular}{@{}l@{\hspace{4pt}}l@{\hspace{2pt}}r@{}}
      \toprule
      \textbf{Field} & \textbf{Description} & \makebox[2pt][r]{\textbf{Data rate [Hz]}}\\
      \midrule
      \texttt{\_time} &
      Timestamp [s] &
      1000\\

      \texttt{/gyro{i}/x,y,z} &
      Body-frame angular rate from IMUs ($i = 0...3$), [rad/s] &
      1000\\

      \texttt{/accel{i}/x,y,z} &
      Body-frame acceleration incl. gravity from IMUs [m/s$^2$] &
      200
      \\

      \texttt{/q\_yrp/yaw,roll,pitch} &
      Estimated robot orientation as yaw, roll, pitch angles [rad] &
      1000\\

      \texttt{/dq\_yrp/*} &
      Time derivatives of yaw, roll, and pitch angles [rad/s] &
      1000\\

      \texttt{/q\_DR/*}, \texttt{/dq\_DR/*}, \texttt{/ddq\_DR/*} &
      Angle,  velocity, accel. of both wheels [rad, rad/s, rad/s$^2$] &
      1000
      \\

      \texttt{/tau\_DR\_command/*} &
      Commanded actuator torques [Nm] &
      167
      \\

      \texttt{/setpoint/*} &
      References for orientation, rates, wheel
      angle and velocities & -- \\

      \texttt{/vicon\_position/*} &
      Global position in world frame from motion capture [m] &
      100
      \\

      \texttt{/vicon\_orientation\_wxyz/*} &
      Robot orientation from motion capture, quaternion [w,x,y,z] &
      100
      \\

      \texttt{battery/voltage} &
      Measured battery voltage [V] & 0.5\\
      \bottomrule
    \end{tabular}
}
\vspace{-1em}
\end{table}

\section{Usage Examples}

\begin{figure}[h!]
    \centering
    \includegraphics[width=\linewidth]{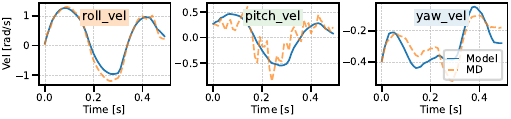}%
    \vspace{-1em}
    \caption{Autoregressive rollout of learned model vs. measured data (MD).}
    \label{fig:model-learning-results}
    \vspace{-1em}
\end{figure}

\fakepar{Dynamics Learning}
We provide an example of dynamics model learning using a multi-layer perceptron (MLP).
The MLP is trained to predict the next state based on the state~$s$, action~$a$, and context~$c$ at time $t$:
$s_{t+1} = \mathrm{MLP}(s_{t},a_{t},c_{t})$.
We use body orientations, angular velocities, wheel positions, and velocities in the state $s$, commanded torques as $a$, and
sub-sample at \SI{100}{\hertz}.
We train on multi-step rollouts, i.e., we roll out the MLP autoregressively for 50 steps starting at $s_0$ from the dataset and then compute a mean squared error loss of the model rollout and the real-world data.
Fig.~\ref{fig:model-learning-results} shows the autoregressive predictions of the final model on a hold-out test trajectory.
\hfill 
\smash{
    \begin{tikzpicture}[baseline=(btn.base)]
        \node(btn) [
            fill=pyblue!10, 
            draw=pyblue!40, 
            rounded corners=1pt, 
            inner sep=2pt,
            blur shadow={ 
                shadow xshift=1pt, 
                shadow yshift=-1pt, 
                shadow blur radius=2pt, 
                shadow opacity=30 
            },
        ] {
            \href{https://github.com/wheelbot/dataset/tree/main/examples/dynamics_model_learning}{
                \scriptsize \textcolor{pyblue}{\faPython} \enspace \textbf{Full Example}
            }
        };
    \end{tikzpicture}
}

\begin{figure}[h!]
    \centering
    \includegraphics[width=\linewidth]{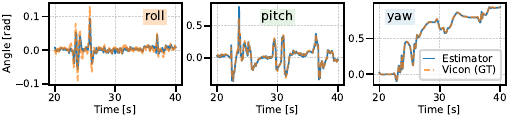}%
    \vspace{-1em}
    \caption{Offline evaluation of an estimator against Vicon ground truth data.}%
    \label{fig:estimator-results}
    \vspace{-1em}
\end{figure}

\fakepar{State Estimation}
Due to the availability of ground truth motion capture data, the dataset can be used to benchmark estimators.
We implement the orientation estimator from~\cite{geist2022wheelbot} as an example in pure Python and compare it with ground-truth from motion capture in Fig.~\ref{fig:estimator-results}.
\hfill 
\smash{
    \begin{tikzpicture}[baseline=(btn2.base)]
        \node(btn2) [
            fill=pyblue!10, 
            draw=pyblue!40, 
            rounded corners=1pt, 
            inner sep=2pt,
            blur shadow={ 
                shadow xshift=1pt, 
                shadow yshift=-1pt, 
                shadow blur radius=2pt, 
                shadow opacity=30 
            },
        ] {
            \href{https://github.com/wheelbot/dataset/tree/main/examples/state_estimation}{
                \scriptsize \textcolor{pyblue}{\faPython} \enspace \textbf{Full Example}
            }
        };
    \end{tikzpicture}
}

\begin{figure}[h!]
    \centering
    \vspace{-0.5em}
    \includegraphics[width=\linewidth]{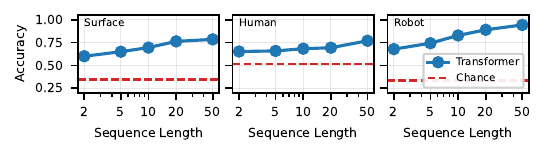}%
    \vspace{-1.5em}
    \caption{Classification accuracies over sequence lengths.}%
    \label{fig:classification}
    \vspace{-1em}
\end{figure}

\fakepar{Time-series Transformer Classifier}
In this example, we train a small time-series transformer classification model that predicts the floor, human or geometric reference, and robot instance from sequences of accelerometer and gyroscope readings.
We report the resulting classification accuracy over the sequence length in Fig.~\ref{fig:classification}.%
\hfill 
\smash{
    \begin{tikzpicture}[baseline=(btn.base)]
        \node(btn) [
            fill=pyblue!10, 
            draw=pyblue!40, 
            rounded corners=1pt, 
            inner sep=2pt,
            blur shadow={ 
                shadow xshift=1pt, 
                shadow yshift=-1pt, 
                shadow blur radius=2pt, 
                shadow opacity=30 
            },
        ] {
            \href{https://github.com/wheelbot/dataset/tree/main/examples/timeseries_classification}{
                \scriptsize \textcolor{pyblue}{\faPython} \enspace \textbf{Full Example}
            }
        };
    \end{tikzpicture}
}

\section{Conclusion and Outlook}
We present a large (\SI{11}{\giga\byte}; 13 mio. state transitions) and diverse dataset of the Mini Wheelbot accompanied by a Python package and example implementations.
We hope to foster reproducibility of results and become a benchmark for learning algorithms targeting fast, unstable, nonlinear dynamics.
We aim to expand the dataset with LiDAR and vision after respective hardware updates to the Mini Wheelbot.\looseness=-1

\newpage
\bibliographystyle{IEEEtran}
\bibliography{references}  %

\end{document}